\documentclass[11pt,a4paper]{article}
\usepackage[hyperref]{acl2019}
\usepackage{times}
\usepackage{latexsym}

\usepackage{graphicx}
\usepackage{tabularx}
\usepackage{booktabs}
\usepackage{multirow}
\usepackage{varwidth}
\usepackage{pbox}

\usepackage{amssymb}
\usepackage{amsmath}
\usepackage{pifont}
\newcommand{\xmark}{\ding{55}}%

\usepackage{url}
\usepackage[normalem]{ulem}
\useunder{\uline}{\ul}{}

\aclfinalcopy 
\def\aclpaperid{1775} 

\title{Language Modelling Makes Sense: Propagating Representations through WordNet for Full-Coverage Word Sense Disambiguation}

\author{Daniel Loureiro, Al\'ipio M\'ario Jorge \\
  LIAAD - INESC TEC \\
  Faculty of Sciences - University of Porto, Portugal \\
  {\tt dloureiro@fc.up.pt, amjorge@fc.up.pt}}

\date{}

\begin{document}
\maketitle
\begin{abstract}
Contextual embeddings represent a new generation of semantic representations learned from Neural Language Modelling (NLM) that addresses the issue of meaning conflation hampering traditional word embeddings. In this work, we show that contextual embeddings can be used to achieve unprecedented gains in Word Sense Disambiguation (WSD) tasks. Our approach focuses on creating sense-level embeddings with full-coverage of WordNet, and without recourse to explicit knowledge of sense distributions or task-specific modelling. As a result, a simple Nearest Neighbors ($k$-NN) method using our representations is able to consistently surpass the performance of previous systems using powerful neural sequencing models. We also analyse the robustness of our approach when ignoring part-of-speech and lemma features, requiring disambiguation against the full sense inventory, and revealing shortcomings to be improved. Finally, we explore applications of our sense embeddings for concept-level analyses of contextual embeddings and their respective NLMs.
\end{abstract}

\section{Introduction} \label{sec:introduction}

Word Sense Disambiguation (WSD) is a core task of Natural Language Processing (NLP) which consists in assigning the correct sense to a word in a given context, and has many potential applications \cite{Navigli2009WordSD}. Despite breakthroughs in distributed semantic representations (i.e. word embeddings), resolving lexical ambiguity has remained a long-standing challenge in the field. Systems using non-distributional features, such as It Makes Sense (IMS, \citealp{Zhong2010ItMS}), remain surprisingly competitive against neural sequence models trained end-to-end. A baseline that simply chooses the most frequent sense (MFS) has also proven to be notoriously difficult to surpass.

Several factors have contributed to this limited progress over the last decade, including lack of standardized evaluation, and restricted amounts of sense annotated corpora. Addressing the evaluation issue, \citet{raganato2017word} has introduced a unified evaluation framework that has already been adopted by the latest works in WSD. Also, even though SemCor \cite{Miller1994UsingAS} still remains the largest manually annotated corpus, supervised methods have successfully used label propagation \cite{Yuan2016SemisupervisedWS}, semantic networks \cite{Vial2018ImprovingTC} and glosses \cite{Luo2018IncorporatingGI} in combination with annotations to advance the state-of-the-art. Meanwhile, task-specific sequence modelling architectures based on BiLSTMs or Seq2Seq \cite{D17-1120} haven't yet proven as advantageous for WSD.

Until recently, the best semantic representations at our disposal, such as word2vec \cite{Mikolov2013DistributedRO} and fastText \cite{Bojanowski2017EnrichingWV}, were bound to word types (i.e. distinct tokens), converging information from different senses into the same representations (e.g. `play song' and `play tennis' share the same representation of `play'). These word embeddings were learned from unsupervised Neural Language Modelling (NLM) trained on fixed-length contexts. However, by recasting the same word types across different sense-inducing contexts, these representations became insensitive to the different senses of polysemous words. \citet{CamachoCollados2018FromWT} refer to this issue as the meaning conflation deficiency and explore it more thoroughly in their work.

Recent improvements to NLM have allowed for learning representations that are context-specific and detached from word types. While word embedding methods reduced NLMs to fixed representations after pretraining, this new generation of contextual embeddings employs the pretrained NLM to infer different representations induced by arbitrarily long contexts. Contextual embeddings have already had a major impact on the field, driving progress on numerous downstream tasks. This success has also motivated a number of iterations on embedding models in a short timespan, from context2vec \cite{Melamud2016context2vecLG}, to GPT \cite{radford2018improving}, ELMo \cite{peters2018deep}, and BERT \cite{bert_naacl}.

Being context-sensitive by design, contextual embeddings are particularly well-suited for WSD. In fact, \citet{Melamud2016context2vecLG} and \citet{peters2018deep} produced contextual embeddings from the SemCor dataset and showed competitive results on \citet{raganato2017word}'s WSD evaluation framework, with a surprisingly simple approach based on Nearest Neighbors ($k$-NN). These results were promising, but those works only produced sense embeddings for the small fraction of WordNet \cite{Fellbaum2000WordNetA} senses covered by SemCor, resorting to the MFS approach for a large number of instances. Lack of high coverage annotations is one of the most pressing issues for supervised WSD approaches \cite{Le2018ADD}.

Our experiments show that the simple $k$-NN w/MFS approach using BERT embeddings suffices to surpass the performance of all previous systems. Most importantly, in this work we introduce a method for generating sense embeddings with full-coverage of WordNet, which further improves results (additional 1.9\% F1) while forgoing MFS fallbacks. To better evaluate the fitness of our sense embeddings, we also analyse their performance without access to lemma or part-of-speech features typically used to restrict candidate senses. Representing sense embeddings in the same space as any contextual embeddings generated from the same pretrained NLM eases introspections of those NLMs, and enables token-level intrinsic evaluations based on $k$-NN WSD performance. We summarize our contributions\footnote{Code and data: \href{https://github.com/danlou/lmms}{github.com/danlou/lmms}} below:
\begin{itemize}
  \item A method for creating sense embeddings for all senses in WordNet, allowing for WSD based on $k$-NN without MFS fallbacks.
  \item Major improvement over the state-of-the-art on cross-domain WSD tasks, while exploring the strengths and weaknesses of our method.
  \item Applications of our sense embeddings for concept-level analyses of NLMs.
\end{itemize}

\section{Language Modelling Representations} \label{sec:lmreps}

Distributional semantic representations learned from Unsupervised Neural Language Modelling (NLM) are currently used for most NLP tasks. In this section we cover aspects of word and contextual embeddings, learned from from NLMs, that are particularly relevant for our work.

\subsection{Static Word Embeddings} \label{sec:static}

Word embeddings are distributional semantic representations usually learned from NLM under one of two possible objectives: predict context words given a target word (Skip-Gram), or the inverse (CBOW) (word2vec, \citealp{Mikolov2013DistributedRO}). In both cases, context corresponds to a fixed-length window sliding over tokenized text, with the target word at the center. These modelling objectives are enough to produce dense vector-based representations of words that are widely used as powerful initializations on neural modelling architectures for NLP. As we explained in the introduction, word embeddings are limited by meaning conflation around word types, and reduce NLM to fixed representations that are insensitive to contexts. However, with fastText \cite{Bojanowski2017EnrichingWV} we're not restricted to a finite set of representations and can compositionally derive representations for word types unseen during training.

\subsection{Contextual Embeddings} \label{sec:contextual}

The key differentiation of contextual embeddings is that they are context-sensitive, allowing the same word types to be represented differently according to the contexts in which they occurr. In order to be able to produce new representations induced by different contexts, contextual embeddings employ the pretrained NLM for inferences. Also, the NLM objective for contextual embeddings is usually directional, predicting the previous and/or next tokens in arbitrarily long contexts (usually sentences). ELMo \cite{peters2018deep} was the first implementation of contextual embeddings to gain wide adoption, but it was shortly after followed by BERT \cite{bert_naacl} which achieved new state-of-art results on 11 NLP tasks. Interestingly, BERT's impressive results were obtained from task-specific fine-tuning of pretrained NLMs, instead of using them as features in more complex models, emphasizing the quality of these representations.

\section{Word Sense Disambiguation (WSD)} \label{sec:wsd}

\noindent There are several lines of research exploring different approaches for WSD \cite{Navigli2009WordSD}. Supervised methods have traditionally performed best, though this distinction is becoming increasingly blurred as works in supervised WSD start exploiting resources used by knowledge-based approaches (e.g. \citealp{luo2018leveraging,Vial2018ImprovingTC}). We relate our work to the best-performing WSD methods, regardless of approach, as well as methods that may not perform as well but involve producing sense embeddings. In this section we introduce the components and related works that are most relevant for our approach.

\subsection{Sense Inventory, Attributes and Relations} \label{sec:wordnet}

The most popular sense inventory is WordNet, a semantic network of general domain concepts linked by a few relations, such as synonymy and hypernymy. WordNet is organized at different abstraction levels, which we describe below. Following the notation used in related works, we represent the main structure of WordNet, called synset, with $lemma_{POS}^{\#}$, where $lemma$ corresponds to the canonical form of a word, $_{POS}$ corresponds to the sense's part-of-speech (\underline{n}oun, \underline{v}erb, \underline{a}djective or adve\underline{r}b), and $^{\#}$ further specifies this entry.

\begin{itemize}
  \item Synsets: groups of synonymous words that correspond to the same sense, e.g. $dog_n^{1}$. 
  \item Lemmas: canonical forms of words, may belong to multiple synsets, e.g. \textit{dog} is a lemma for $dog_n^{1}$ and $chase_v^{1}$, among others.
  \item Senses: lemmas specifed by sense (i.e. sensekeys), e.g. \textit{{dog\small{\%1:05:00::}}}, and \textit{{domestic\_dog\small{\%1:05:00::}}} are senses of $dog_n^{1}$.
\end{itemize}

Each synset has a number of attributes, of which the most relevant for this work are:

\begin{itemize}
  \item Glosses: dictionary definitions, e.g. $dog_n^{1}$ has the definition `a member of the genus Ca...'.
  \item Hypernyms: `type of' relations between synsets, e.g. $dog_n^{1}$ is a hypernym of $pug_n^{1}$.
  \item Lexnames: syntactical and logical groupings, e.g. the lexname for $dog_n^{1}$ is \textit{noun.animal}.
\end{itemize}

In this work we're using WordNet 3.0, which contains 117,659 synsets, 206,949 unique senses, 147,306 lemmas, and 45 lexnames.

\subsection{WSD State-of-the-Art} \label{sec:sota}

While non-distributional methods, such as \citet{Zhong2010ItMS}'s IMS, still perform competitively, there are have been several noteworthy advancements in the last decade using distributional representations from NLMs. \citet{iacobacci2016embeddings} improved on IMS's performance by introducing word embeddings as additional features.

\citet{Yuan2016SemisupervisedWS} achieved significantly improved results by leveraging massive corpora to train a NLM based on an LSTM architecture. This work is contemporaneous with \citet{Melamud2016context2vecLG}, and also uses a very similar approach for generating sense embeddings and relying on $k$-NN w/MFS for predictions. Although most performance gains stemmed from their powerful NLM, they also introduced a label propagation method that further improved results in some cases. Curiously, the objective \citet{Yuan2016SemisupervisedWS} used for NLM (predicting held-out words) is very evocative of the cloze-style Masked Language Model introduced by \citet{bert_naacl}. \citet{Le2018ADD} replicated this work and offers additional insights.

\citet{D17-1120} trained neural sequencing models for end-to-end WSD. This work reframes WSD as a translation task where sequences of words are translated into sequences of senses. The best result was obtained with a BiLSTM trained with auxilliary losses specific to parts-of-speech and lexnames. Despite the sophisticated modelling architecture, it still performed on par with \citet{iacobacci2016embeddings}.

The works of \citet{Melamud2016context2vecLG} and \citet{peters2018deep} using contextual embeddings for WSD showed the potential of these representations, but still performed comparably to IMS.

Addressing the issue of scarce annotations, recent works have proposed methods for using resources from knowledge-based approaches. \citet{luo2018leveraging} and \citet{Luo2018IncorporatingGI} combine information from glosses present in WordNet, with NLMs based on BiLSTMs, through memory networks and co-attention mechanisms, respectively. \citet{Vial2018ImprovingTC} follows \citet{D17-1120}'s BiLSTM method, but leverages the semantic network to strategically reduce the set of senses required for disambiguating words.

All of these works rely on MFS fallback. Additionally, to our knowledge, all also perform disambiguation only against the set of admissible senses given the word's lemma and part-of-speech.

\subsection{Other methods with Sense Embeddings} \label{sec:senseembeddings}

Some works may no longer be competitive with the state-of-the-art, but nevertheless remain relevant for the development of sense embeddings. We recommend the recent survey of \citet{CamachoCollados2018FromWT} for a thorough overview of this topic, and highlight a few of the most relevant methods. \citet{Chen2014AUM} initializes sense embeddings using glosses and adapts the Skip-Gram objective of word2vec to learn and improve sense embeddings jointly with word embeddings. \citet{Rothe2015AutoExtendEW}'s AutoExtend method uses pretrained word2vec embeddings to compose sense embeddings from sets of synonymous words. \citet{CamachoCollados2016NasariIE} creates the NASARI sense embeddings using structural knowledge from large multilingual semantic networks.

These methods represent sense embeddings in the same space as the pretrained word embeddings, however, being based on fixed embedding spaces, they are much more limited in their ability to generate contextual representations to match against. Furthermore, none of these methods (or those in \S \ref{sec:sota}) achieve full-coverage of the +200K senses in WordNet.

\section{Method} \label{sec:method}

\begin{figure}[htb]
  \centering
  \includegraphics[width=0.4\textwidth]{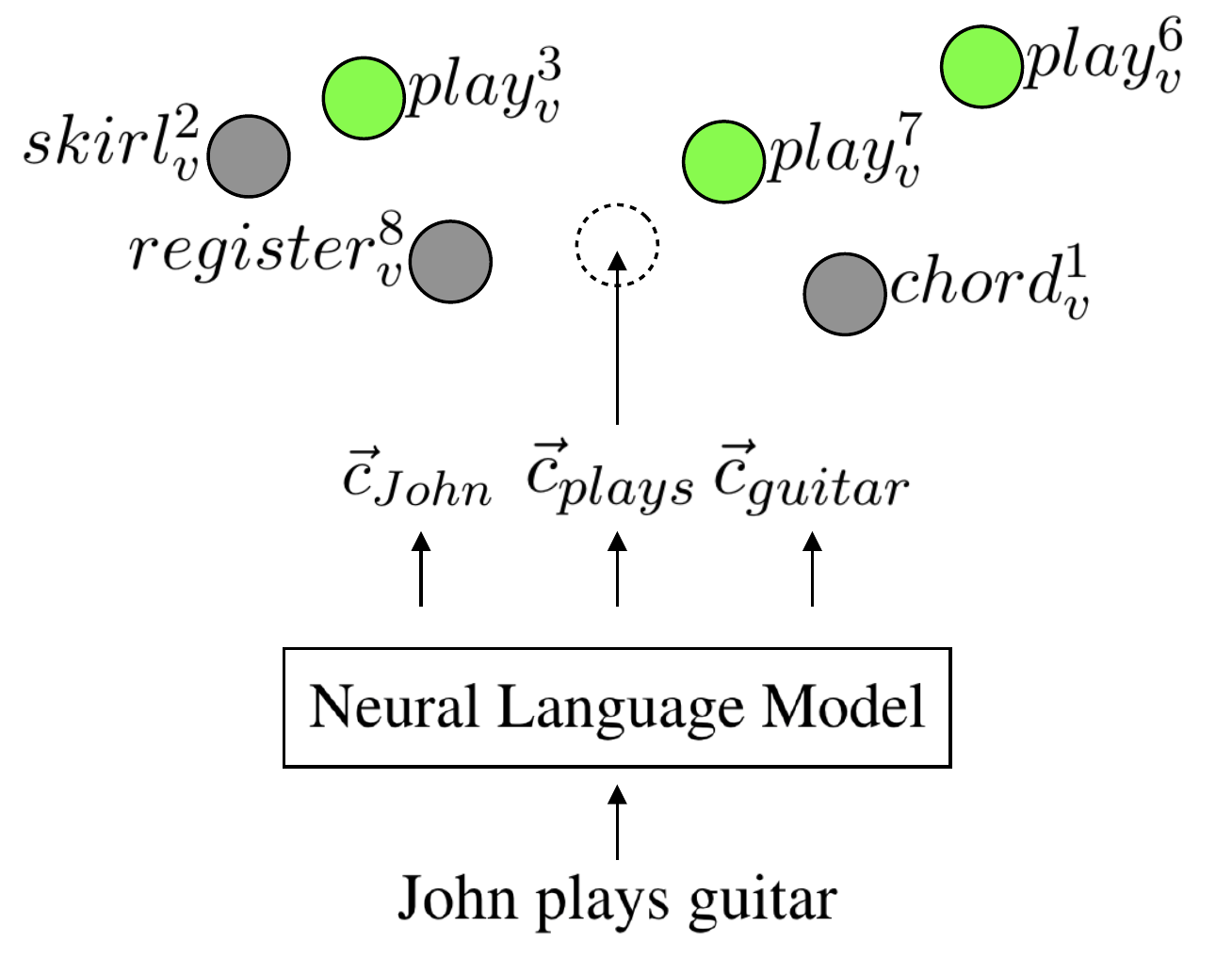}
  \caption{Illustration of our $k$-NN approach for WSD, which relies on full-coverage sense embeddings represented in the same space as contextualized embeddings. For simplification, we label senses as synsets. Grey nodes belong to different lemmas (see \S \ref{sec:usm}).}
  \label{fig:knn}
\end{figure}

Our WSD approach is strictly based on $k$-NN (see Figure \ref{fig:knn}), unlike any of the works referred previously. We avoid relying on MFS for lemmas that do not occur in annotated corpora by generating sense embeddings with full-coverage of WordNet. Our method starts by generating sense embeddings from annotations, as done by other works, and then introduces several enhancements towards full-coverage, better performance and increased robustness. In this section, we cover each of these techniques.

\subsection{Embeddings from Annotations} \label{sec:annotations}

Our set of full-coverage sense embeddings is bootstrapped from sense-annotated corpora. Sentences containing sense-annotated tokens (or spans) are processed by a NLM in order to obtain contextual embeddings for those tokens. After collecting all sense-labeled contextual embeddings, each sense  embedding is determined by averaging its corresponding contextual embeddings. Formally, given $n$ contextual embeddings $\vec{c}$ for some sense $s$:

$$\vec{v}_{s} = \frac{1}{n}\sum_{i=1}^{n}\vec{c}_i , dim(\vec{v}_{s}) = 1024$$

In this work we use pretrained ELMo and BERT models to generate contextual embeddings. These models can be identified and replicated with the following details:

\begin{itemize}
  \item ELMo: 1024 (2x512) embedding dimensions, 93.6M parameters. Embeddings from top layer (2).
  \item BERT: 1024 embedding dimensions, 340M parameters, cased. Embeddings from sum of top 4 layers ([-1,-4])\footnote{This was the configuration that performed best out of the ones on Table 7 of \citet{devlin2018bert}.}.
\end{itemize}

BERT uses WordPiece tokenization that doesn't always map to token-level annotations (e.g. `multiplication' becomes `multi', `\#\#plication'). We use the average of subtoken embeddings as the token-level embedding. Unless specified otherwise, our LMMS method uses BERT.

\subsection{Extending Annotation Coverage} \label{sec:extend}

As many have emphasized before \cite{Navigli2009WordSD,CamachoCollados2018FromWT,Le2018ADD}, the lack of sense annotations is a major limitation of supervised approaches for WSD. We address this issue by taking advantage of the semantic relations in WordNet to extend the annotated signal to other senses. Semantic networks are often explored by knowledge-based approaches, and some recent works in supervised approaches as well \cite{luo2018leveraging,Vial2018ImprovingTC}. The guiding principle behind these approaches is that sense-level representations can be imputed (or improved) from other representations that are known to correspond to generalizations due to the network's taxonomical structure. \citet{Vial2018ImprovingTC} leverages relations in WordNet to reduce the sense inventory to a minimal set of entries, making the task easier to model while maintaining the ability to distinguish senses. We take the inverse path of leveraging relations to produce representations for additional senses.

On \S \ref{sec:wordnet} we covered synsets, hypernyms and lexnames, which correspond to increasingly abstract generalizations. Missing sense embeddings are imputed from the aggregation of sense embeddings at each of these abstraction levels. In order to get embeddings that are representative of higher-level abstractions, we simply average the embeddings of all lower-level constituents. Thus, a synset embedding corresponds to the average of all of its sense embeddings, a hypernym embedding corresponds to the average of all of its synset embeddings, and a lexname embedding corresponds to the average of a larger set of synset embeddings. All lower abstraction representations are created before next-level abstractions to ensure that higher abstractions make use of lower generalizations. More formally, given all missing senses in WordNet $\hat{s} \in {W}$, their synset-specific sense embeddings $S_{\hat{s}}$, hypernym-specific synset embeddings $H_{\hat{s}}$, and lexname-specific synset embeddings $L_{\hat{s}}$, the procedure has the following stages:
$$
\begin{matrix}
(1) & if |S_{\hat{s}}| > 0 , & \vec{v}_{\hat{s}} = \frac{1}{|S_{\hat{s}}|}\sum\vec{v}_{s} , \forall \vec{v}_{s} \in S_{\hat{s}} \\\\

(2) & if |H_{\hat{s}}| > 0 , & \vec{v}_{\hat{s}} = \frac{1}{|H_{\hat{s}}|}\sum\vec{v}_{syn} , \forall \vec{v}_{syn} \in H_{\hat{s}} \\\\

(3) & if |L_{\hat{s}}| > 0 , & \vec{v}_{\hat{s}} = \frac{1}{|L_{\hat{s}}|}\sum\vec{v}_{syn} , \forall \vec{v}_{syn} \in L_{\hat{s}}
\end{matrix}
$$

In Table \ref{tab:extension} we show how much coverage extends while improving both recall and precision.
\begin{table}[h]
\centering
\resizebox{0.48\textwidth}{!}{%
\begin{tabular}{@{}lccc@{}}
\toprule
\textbf{} & \textbf{} & \multicolumn{2}{c}{\small{\textbf{F1 / P / R (without MFS)}}} \\ \cmidrule(l){3-4}
\multicolumn{1}{c}{\textbf{Source}} & \textbf{Coverage} & \small{\textbf{BERT}} & \small{\textbf{ELMo}} \\ \midrule \midrule
\multicolumn{1}{l}{SemCor} & 16.11\% & 68.9 / 72.4 / 65.7 & 63.0 / 66.2 / 60.1 \\ \midrule
\multicolumn{1}{l}{+ synset} & 26.97\% & 70.0 / 72.6 / 70.0 & 63.9 / 66.3 / 61.7 \\ \midrule
\multicolumn{1}{l}{+ hypernym} & 74.70\% & 73.0 / 73.6 / 72.4 & 67.2 / 67.7 / 66.6 \\ \midrule
\multicolumn{1}{l}{+ lexname} & 100\% & 73.8 / 73.8 / 73.8 & 68.1 / 68.1 / 68.1 \\
\bottomrule
\end{tabular}
}
\caption{Coverage of WordNet when extending to increasingly abstract representations along with performance on the ALL test set of \citet{raganato2017word}.}
\label{tab:extension}
\end{table}

\subsection{Improving Senses using the Dictionary} \label{sec:dictionary}

There's a long tradition of using glosses for WSD, perhaps starting with the popular work of \citet{Lesk1986AutomaticSD}, which has since been adapted to use distributional representations \cite{Basile2014AnEL}. As a sequence of words, the information contained in glosses can be easily represented in semantic spaces through approaches used for generating sentence embeddings. There are many methods for generating sentence embeddings, but it's been shown that a simple weighted average of word embeddings performs well \cite{Arora2017ASB}.

Our contextual embeddings are produced from NLMs using attention mechanisms, assigning more importance to some tokens over others, so they already come `pre-weighted' and we embed glosses simply as the average of all of their contextual embeddings (without preprocessing). We've also found that introducing synset lemmas alongside the words in the gloss helps induce better contextualized embeddings (specially when glosses are short). Finally, we make our dictionary embeddings ($\vec{v}_d$) sense-specific, rather than synset-specific, by repeating the lemma that's specific to the sense, alongside the synset's lemmas and gloss words. The result is a sense-level embedding, determined without annotations, that is represented in the same space as the sense embeddings we described in the previous section, and can be trivially combined through concatenation or average for improved performance (see Table \ref{tab:ablation}).

Our empirical results show improved performance by concatenation, which we attribute to preserving complementary information from glosses. Both averaging and concatenating representations (previously $L_2$ normalized) also serves to smooth possible biases that may have been learned from the SemCor annotations. Note that while concatenation effectively doubles the size of our embeddings, this doesn't equal doubling the expressiveness of the distributional space, since they're two representations from the same NLM. This property also allows us to make predictions for contextual embeddings (from the same NLM) by simply repeating those embeddings twice, aligning contextual features against sense and dictionary features when computing cosine similarity. Thus, our sense embeddings become:

$$
\vec{v}_s = \begin{bmatrix}
  ||\vec{v}_s||_2\\
  ||\vec{v}_d||_2
  \end{bmatrix}, dim(\vec{v}_s) = 2048
$$

\begin{table*}[htb]
  \centering
  \resizebox{0.85\textwidth}{!}{%
  \begin{tabular}{@{}c|c|c|c|c|c|c|c@{}}
  \toprule
  \textbf{Configurations} & \textbf{LMMS$_{1024}$} & \hspace{1.5cm} & \hspace{1.5cm} & \textbf{LMMS$_{2048}$} & \hspace{1.5cm} & \hspace{1.5cm} & \textbf{LMMS$_{2348}$} \\ \hline \hline
  \textbf{Embeddings} & \textbf{} & \textbf{} & \textbf{} & \textbf{} & \textbf{} & \textbf{} & \textbf{} \\
  Contextual \small{(d=1024)} & \xmark &  & \xmark & \xmark & \xmark &  & \xmark \\
  Dictionary \small{(d=1024)} &  & \xmark & \xmark & \xmark &  & \xmark & \xmark \\
  Static \small{(d=300)} &  &  &  &  & \xmark & \xmark & \xmark \\
  \textbf{Operation} & \textbf{} & \textbf{} & \textbf{} & \textbf{} & \textbf{} & \textbf{} & \textbf{} \\
  Average &  &  & \xmark &  &  &  &  \\
  Concatenation &  &  &  & \xmark & \xmark & \xmark & \xmark \\ \hline
  \textbf{Perf. (F1 on ALL)} & \textbf{} & \textbf{} & \textbf{} & \textbf{} & \textbf{} & \textbf{} & \textbf{} \\
  Lemma \& POS & 73.8 & 58.7 & 75.0 & \textbf{75.4} & 73.9 & 58.7 & \textbf{75.4} \\
  Token \small{(Uninformed)} & 42.7 & 6.1 & 36.5 & 35.1 & 64.4 & 45.0 & \textbf{66.0} \\
  \bottomrule
  \end{tabular}
  }
  \caption{Overview of the different performance of various setups regarding choice of embeddings and combination strategy. All results are for the 1-NN approach on the ALL test set of \citet{raganato2017word}. We also show results that ignore the lemma and part-of-speech features of the test sets to show that the inclusion of static embeddings makes the method significantly more robust to real-world scenarios where such gold features may not be available.}
  \label{tab:ablation}
\end{table*}

\subsection{Morphological Robustness} \label{sec:morphology}

WSD is expected to be performed only against the set of candidate senses that are specific to a target word's lemma. However, as we'll explain in \S \ref{sec:usm}, there are cases where it's undesirable to restrict the WSD process.

We leverage word embeddings specialized for morphological representations to make our sense embeddings more resilient to the absence of lemma features, achieving increased robustness. This addresses a problem arising from the susceptibility of contextual embeddings to become entirely detached from the morphology of their corresponding tokens, due to interactions with other tokens in the sentence.

We choose fastText \cite{Bojanowski2017EnrichingWV} embeddings (pretrained on CommonCrawl), which are biased towards morphology, and avoid Out-of-Vocabulary issues as explained in \S \ref{sec:static}. We use fastText to generate static word embeddings for the lemmas ($\vec{v}_l$) corresponding to all senses, and concatenate these word embeddings to our previous embeddings. When making predictions, we also compute fastText embeddings for tokens, allowing for the same alignment explained in the previous section. This technique effectively makes sense embeddings of morphologically related lemmas more similar. Empirical results (see Table \ref{tab:ablation}) show that introducing these static embeddings is crucial for achieving satisfactory performance when not filtering candidate senses.

\noindent Our final, most robust, sense embeddings are thus:

$$
\vec{v}_s = \begin{bmatrix}
  ||\vec{v}_s||_2 \\
  ||\vec{v}_d||_2 \\
  ||\vec{v}_l||_2  
  \end{bmatrix}, dim(\vec{v}_s) = 2348
$$

\section{Experiments}  \label{sec:experiments}

Our experiments centered on evaluating our solution on \citet{raganato2017word}'s set of cross-domain WSD tasks. In this section we compare our results to the current state-of-the-art, and provide results for our solution when disambiguating against the full set of possible senses in WordNet, revealing shortcomings to be improved.

\begin{table*}[htb]
\centering
\resizebox{0.85\textwidth}{!}{%
\begin{tabular}{@{}ccccccc@{}}
\toprule
\multirow{2}{*}{\textbf{Model}} & \textbf{Senseval2} & \textbf{Senseval3} & \textbf{SemEval2007} & \textbf{SemEval2013} & \textbf{SemEval2015} & \textbf{ALL} \\
& \small{(n=2,282)} & \small{(n=1,850)} & \small{(n=455)} & \small{(n=1,644)} & \small{(n=1,022)} & \small{(n=7,253)} \\ \midrule \midrule
MFS$^\dagger$ (Most Frequent Sense) & {65.6} & {66.0} & {54.5} & {63.8} & {67.1} & {64.8} \\
IMS$^\dagger$ \citeyearpar{Zhong2010ItMS} & {70.9} & {69.3} & {61.3} & {65.3} & {69.5} & {68.4} \\
IMS + embeddings$^\dagger$ \citeyearpar{iacobacci2016embeddings} & {72.2} & {70.4} & {62.6} & {65.9} & {71.5} & {69.6} \\
context2vec $k$-NN$^\dagger$ \citeyearpar{Melamud2016context2vecLG} & {71.8} & {69.1} & {61.3} & {65.6} & {71.9} & {69.0} \\
word2vec $k$-NN \citeyearpar{Yuan2016SemisupervisedWS} & {67.8} & {62.1} & {58.5} & {66.1} & {66.7} & {-} \\
LSTM-LP (Label Prop.) \citeyearpar{Yuan2016SemisupervisedWS} & {\underline{73.8}} & {\underline{71.8}} & {\underline{63.5}} & {69.5} & {72.6} & {-} \\
Seq2Seq (Task Modelling) \citeyearpar{D17-1120} & {70.1} & {68.5} & {63.1*} & {66.5} & {69.2} & {68.6*} \\
BiLSTM (Task Modelling) \citeyearpar{D17-1120} & {72.0} & {69.1} & {64.8*} & {66.9} & {71.5} & {69.9*} \\
ELMo $k$-NN \citeyearpar{peters2018deep} & {71.5} & {67.5} & {57.1} & {65.3} & {69.9} & {67.9} \\
HCAN (Hier. Co-Attention) \citeyearpar{luo2018leveraging} & {72.8} & {70.3} & {-*} & {68.5} & {\underline{72.8}} & {-*} \\
BiLSTM w/Vocab. Reduction \citeyearpar{Vial2018ImprovingTC} & {72.6} & {70.4} & {61.5} & {\underline{70.8}} & {71.3} & {70.8} \\ \midrule
BERT $k$-NN & {\textbf{76.3}} & {73.2} & {66.2} & {71.7} & {74.1} & {73.5} \\
LMMS$_{2348}$ (ELMo) & {68.1} & {64.7} & {53.8} & {66.9} & {69.0} & {66.2} \\
LMMS$_{2348}$ (BERT) & {\textbf{76.3}} & {\textbf{75.6}} & {\textbf{68.1}} & {\textbf{75.1}} & {\textbf{77.0}} & {\textbf{75.4}} \\
\bottomrule
\end{tabular}%
}
\caption{Comparison with other works on the test sets of \citet{raganato2017word}. All works used sense annotations from SemCor as supervision, although often different pretrained embeddings. $^\dagger$ - reproduced from \citet{raganato2017word}; * - used as a development set; bold - new state-of-the-art (SOTA); underlined - previous SOTA.}
\label{tab:eval}
\end{table*}
  
\subsection{All-Words Disambiguation}  \label{sec:eval}

In Table \ref{tab:eval} we show our results for all tasks of \citet{raganato2017word}'s evaluation framework. We used the framework's scoring scripts to avoid any discrepancies in the scoring methodology. Note that the $k$-NN referred in Table \ref{tab:eval} always refers to the closest neighbor, and relies on MFS fallbacks.

The first noteworthy result we obtained was that simply replicating \citet{peters2018deep}'s method for WSD using BERT instead of ELMo, we were able to significantly, and consistently, surpass the performance of all previous works. When using our method (LMMS), performance still improves significantly over the previous impressive results (+1.9 F1 on ALL, +3.4 F1 on SemEval 2013). Interestingly, we found that our method using ELMo embeddings didn't outperform ELMo $k$-NN with MFS fallback, suggesting that it's necessary to achieve a minimum competence level of embeddings from sense annotations (and glosses) before the inferred sense embeddings become more useful than MFS.

\begin{figure}[htb]
  \centering
  \includegraphics[width=0.35\textwidth]{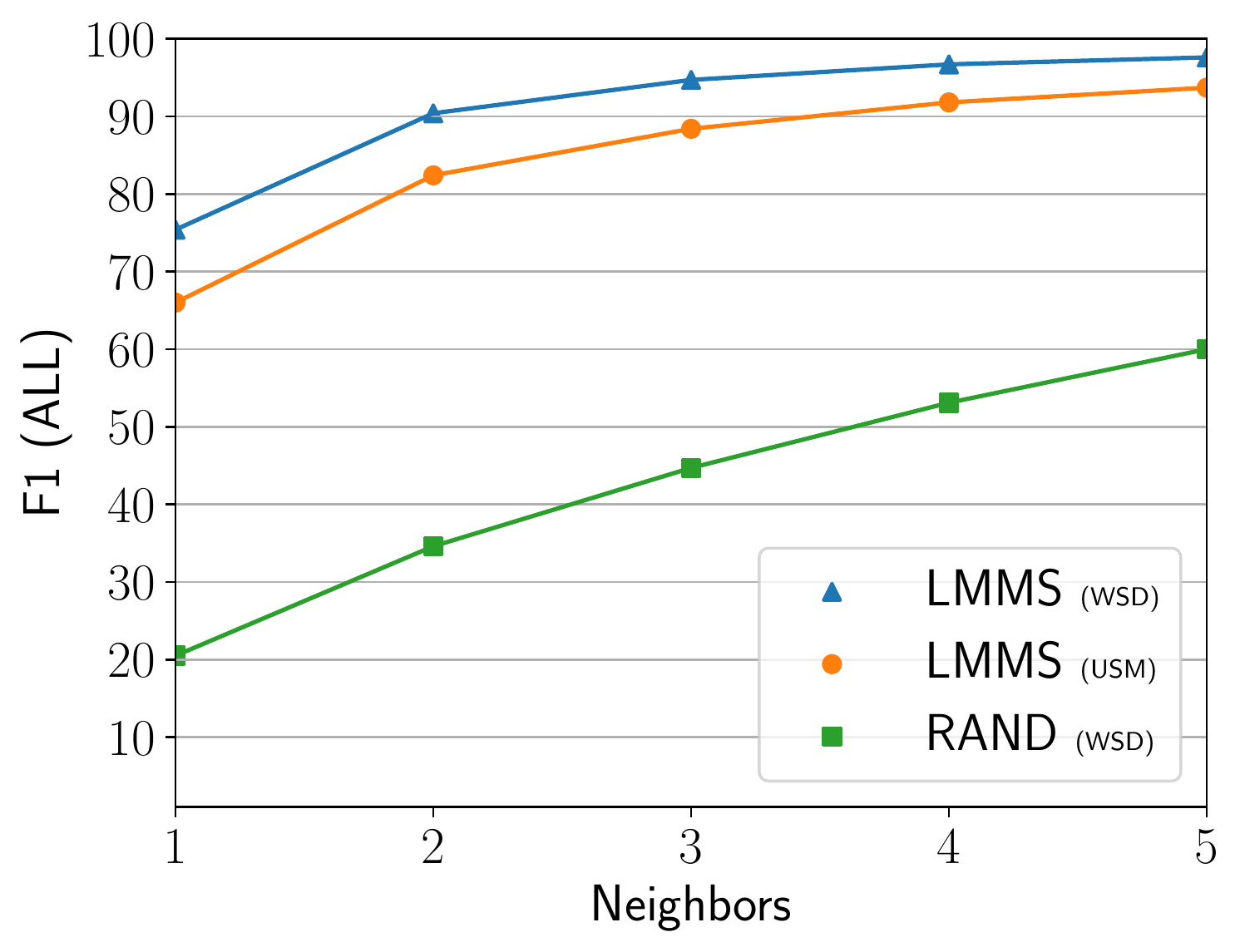}
  \caption{Performance gains with LMMS$_{2348}$ when accepting additional neighbors as valid predictions.}
  \label{fig:knn_add}
\end{figure}

In Figure \ref{fig:knn_add} we show results when considering additional neighbors as valid predictions, together with a random baseline considering that some target words may have less senses than the number of accepted neighbors (always correct).

\subsection{Part-of-Speech Mismatches}  \label{sec:pos}

The solution we introduced in \S \ref{sec:morphology} addressed missing lemmas, but we didn't propose a solution that addressed missing POS information. Indeed, the confusion matrix in Table \ref{tab:confusion} shows that a large number of target words corresponding to verbs are wrongly assigned senses that correspond to adjectives or nouns. We believe this result can help motivate the design of new NLM tasks that are more capable of distinguishing between verbs and non-verbs.

\begin{table}[htb]
  \centering
  \resizebox{0.40\textwidth}{!}{%
  \renewcommand{\arraystretch}{1.5}
  \begin{tabular}{@{}c||cccc@{}}
  \toprule
  \textbf{WN-POS} & \textbf{NOUN} & \textbf{VERB} & \textbf{ADJ} & \textbf{ADV} \\ \hline \hline
  \textbf{NOUN} & 96.95\% & 1.86\% & 0.86\% & 0.33\% \\ \hline
  \textbf{VERB} & {\ul 9.08\%} & 70.82\% & {\ul 19.98\%} & 0.12\% \\ \hline
  \textbf{ADJ} & {\ul 4.50\%} & 0\% & 92.27\% & 2.93\% \\ \hline
  \textbf{ADV} & 2.02\% & 0.29\% & 2.60\% & 95.09\% \\
  \bottomrule
  \end{tabular}
  \renewcommand{\arraystretch}{1}
  }
  \caption{POS Confusion Matrix for Uninformed Sense Matching on the ALL testset using LMMS$_{2348}$.}
  \label{tab:confusion}
\end{table}

\subsection{Uninformed Sense Matching} \label{sec:usm}
WSD tasks are usually accompanied by auxilliary parts-of-speech (POSs) and lemma features for restricting the number of possible senses to those that are specific to a given lemma and POS. Even if those features aren't provided (e.g. real-world applications), it's sensible to use lemmatizers or POS taggers to extract them for use in WSD. However, as is the case with using MFS fallbacks, this filtering step obscures the true impact of NLM representations on $k$-NN solutions.

Consequently, we introduce a variation on WSD, called Uninformed Sense Matching (USM), where disambiguation is always performed against the full set of sense embeddings (i.e. +200K vs. a maximum of 59). This change makes the task much harder (results on Table \ref{tab:ablation}), but offers some insights into NLMs, which we cover briefly in \S \ref{sec:knowledge}.

\subsection{Use of World Knowledge} \label{sec:knowledge}

It's well known that WSD relies on various types of knowledge, including commonsense and selectional preferences \cite{Lenat1986CYCUC,Resnik1997SelectionalPA}, for example. Using our sense embeddings for Uninformed Sense Matching allows us to glimpse into how NLMs may be interpreting  contextual information with regards to the knowledge represented in WordNet. In Table \ref{tab:knowledge} we show a few examples of senses matched at the token-level, suggesting that entities were topically understood and this information was useful to disambiguate verbs. These results would be less conclusive without full-coverage of WordNet.

\begin{table*}[htb]
  \centering
  \resizebox{0.90\textwidth}{!}{%
  \begin{tabular}{@{}p{3cm}p{3cm}p{3cm}p{3cm}p{3cm}p{3cm}@{}}
  \textbf{Marlon$^\star$} & \textbf{Brando$^\star$} & {\ul \textbf{played}} & \textbf{Corleone$^\star$} & \textbf{in} & \textbf{Godfather$^\star$} \\
  \small{$person_{n}^{1}$} & \small{$person_{n}^{1}$} & \small{$act_{v}^{3}$} & \small{$syndicate_{n}^{1}$} & \small{$movie_{n}^{1}$} & \small{$location_{n}^{1}$} \\
  \small{$womanizer_{n}^{1}$} & \small{$group_{n}^{1}$} & \small{$make_{v}^{42}$} & \small{$mafia_{n}^{1}$} & \small{$telefilm_{n}^{1}$} & \small{$here_{n}^{1}$} \\
  \small{$bustle_{n}^{1}$} & \small{$location_{n}^{1}$} & \small{$emote_{v}^{1}$} & \small{$person_{n}^{1}$} & \small{$final$\_$cut_{n}^{1}$} & \small{$there_{n}^{1}$} \\ \midrule
  \pbox{1.2\textwidth}{$\mathbf{act_{v}^{3}}$: play a role or part; $\mathbf{make_{v}^{42}}$: represent fictiously, as in a play, or pretend to be or act like; $\mathbf{emote_{v}^{1}}$: give expression or emotion to, in a stage or movie role.} \\
  &  &  &  &  &  \\
  \textbf{Serena$^\star$} & \textbf{Williams} & {\ul \textbf{played}} & \textbf{Kerber$^\star$} & \textbf{in} & \textbf{Wimbledon$^\star$} \\
  \small{$person_{n}^{1}$} & \small{$professional$\_$tennis_{n}^{1}$} & \small{$play_{v}^{1}$} & \small{$person_{n}^{1}$} & \small{$win_{v}^{1}$} & \small{$tournament_{n}^{1}$} \\
  \small{$therefore_{r}^{1}$} & \small{$tennis_{n}^{1}$} & \small{$line$\_$up_{v}^{6}$} & \small{$group_{n}^{1}$} & \small{$romp_{v}^{3}$} & \small{$world$\_$cup_{n}^{1}$} \\
  \small{$reef_{n}^{1}$} & \small{$singles_{n}^{1}$} & \small{$curl_{v}^{5}$} & \small{$take$\_$orders_{v}^{2}$} & \small{$carry_{v}^{38}$} & \small{$elimination$\_$tournament_{n}^{1}$} \\ \midrule
  \pbox{1.2\textwidth}{$\mathbf{play_{v}^{1}}$: participate in games or sport; $\mathbf{line}$\_$\mathbf{up_{v}^{6}}$: take one's position before a kick-off; $\mathbf{curl_{v}^{5}}$: play the Scottish game of curling.} \\
  &  &  &  &  &  \\
  \textbf{David} & \textbf{Bowie$^\star$} & {\ul \textbf{played}} & \textbf{Warszawa$^\star$} & \textbf{in} & \textbf{Tokyo} \\
  \small{$person_{n}^{1}$} & \small{$person_{n}^{1}$} & \small{$play_{v}^{14}$} & \small{$poland_{n}^{1}$} & \small{$originate$\_$in_{n}^{1}$} & \small{$tokyo_{n}^{1}$} \\
  \small{$amati_{n}^{2}$} & \small{$folk$\_$song_{n}^{1}$} & \small{$play_{v}^{6}$} & \small{$location_{n}^{1}$} & \small{$in_{r}^{1}$} & \small{$japan_{n}^{1}$} \\
  \small{$guarnerius_{n}^{3}$} & \small{$fado_{n}^{1}$} & \small{$riff_{v}^{2}$} & \small{$here_{n}^{1}$} & \small{$take$\_$the$\_$field_{v}^{2}$} & \small{$japanese_{a}^{1}$} \\ \midrule
  \pbox{1.2\textwidth}{$\mathbf{play_{v}^{14}}$: perform on a certain location; $\mathbf{play_{v}^{6}}$: replay (as a melody); $\mathbf{riff_{v}^{2}}$: play riffs.} \\
  \end{tabular}%
  }
  \caption{Examples controlled for syntactical changes to show how the correct sense for `played' can be induced accordingly with the mentioned entities, suggesting that disambiguation is supported by world knowledge learned during LM pretraining. Words with $^\star$ never occurred in SemCor. Senses shown correspond to the top 3 matches in LMMS$_{1024}$ for each token's contextual embedding (uninformed). For clarification, below each set of matches are the WordNet definitions for the top disambiguated senses of `played'.}
  \label{tab:knowledge}
\end{table*}

\section{Other Applications}  \label{sec:otherapps}

\begin{figure}[htb]
  \centering
  \includegraphics[width=0.49\textwidth]{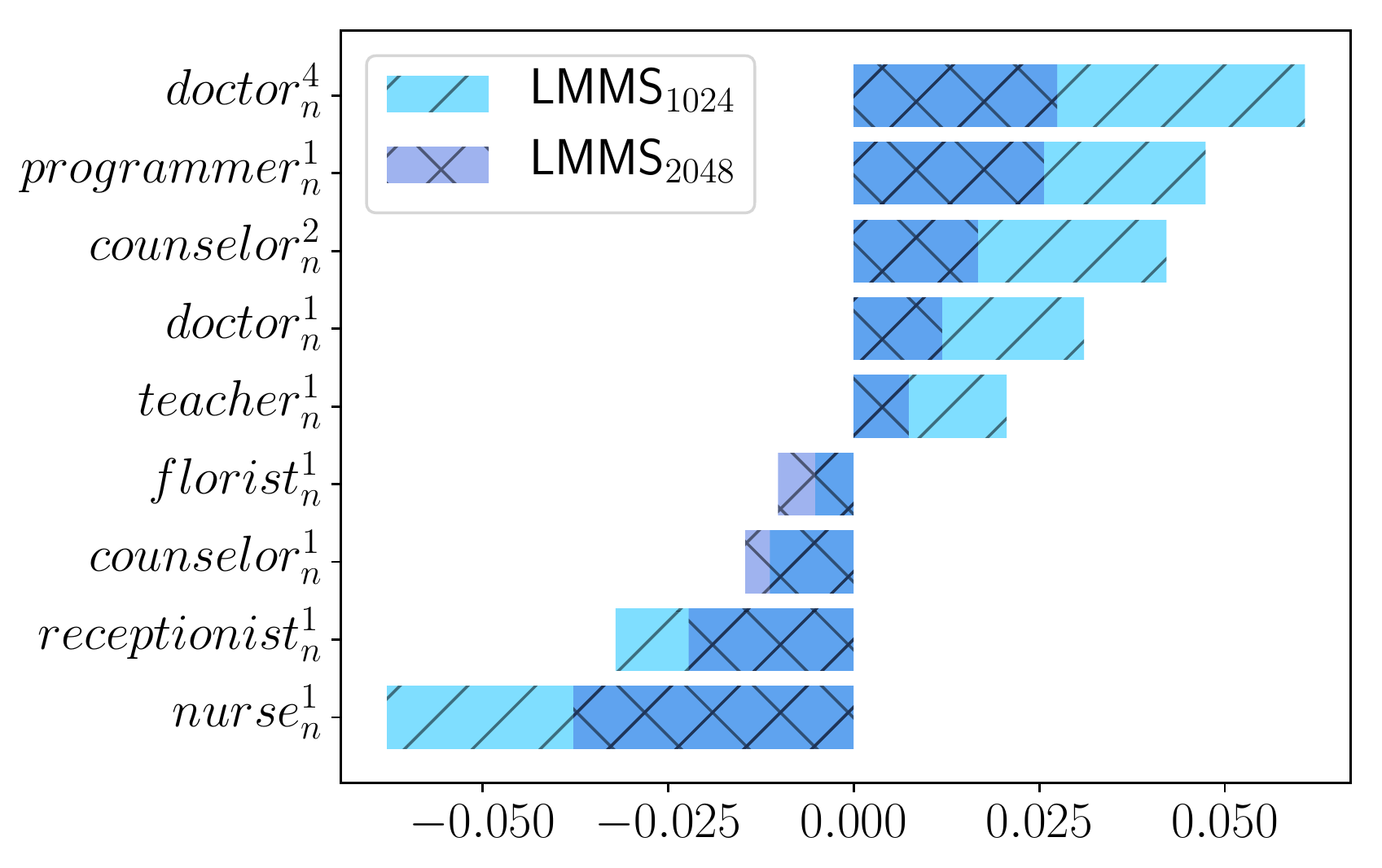}
  \caption{Examples of gender bias found in the sense vectors. Positive values quantify bias towards $man_{n}^{1}$, while negative values quantify bias towards $woman_{n}^{1}$.}
  \label{fig:gender}
\end{figure}

Analyses of conventional word embeddings have revealed gender or stereotype biases \cite{Bolukbasi2016ManIT,Caliskan2017SemanticsDA} that may have unintended consequences in downstream applications. With contextual embeddings we don't have sets of concept-level representations for performing similar analyses. Word representations can naturally be derived from averaging their contextual embeddings occurring in corpora, but then we're back to the meaning conflation issue described earlier. We believe that our sense embeddings can be used as representations for more easily making such analyses of NLMs. In Figure \ref{fig:gender} we provide an example that showcases meaningful differences in gender bias, including for lemmas shared by different senses ($doctor$: PhD vs. medic, and $counselor$: therapist vs. summer camp supervisor). The bias score for a given synset $s$ was calculated as following:
$$ bias(s) = sim(\vec{v}_{man_{n}^{1}}, \vec{v}_{s}) - sim(\vec{v}_{woman_{n}^{1}}, \vec{v}_{s})$$

Besides concept-level analyses, these sense embeddings can also be useful in applications that don't rely on a particular inventory of senses. In \citet{LoureiroWiC}, we show how similarities between matched sense embeddings and contextual embeddings are used for training a classifier that determines whether a word that occurs in two different sentences shares the same meaning.

\section{Future Work}  \label{sec:future}

In future work we plan to use multilingual resources (i.e. embeddings and glosses) for improving our sense embeddings and evaluating on multilingual WSD. We're also considering exploring a semi-supervised approach where our best embeddings would be employed to automatically annotate corpora, and repeat the process described on this paper until convergence, iteratively fine-tuning sense embeddings.
We expect our sense embeddings to be particularly useful in downstream tasks that may benefit from relational knowledge made accessible through linking words (or spans) to commonsense-level concepts in WordNet, such as Natural Language Inference.

\section{Conclusion}  \label{sec:conclusion}

This paper introduces a method for generating sense embeddings that allows a clear improvement of the current state-of-the-art on cross-domain WSD tasks. We leverage contextual embeddings, semantic networks and glosses to achieve full-coverage of all WordNet senses. Consequently, we're able to perform WSD with a simple $1$-NN, without recourse to MFS fallbacks or task-specific modelling. Furthermore, we introduce a variant on WSD for matching contextual embeddings to all WordNet senses, offering a better understanding of the strengths and weaknesses of representations from NLM. Finally, we explore applications of our sense embeddings beyond WSD, such as gender bias analyses.

\section{Acknowledgements}  \label{sec:acknowledgements}

This work is financed by National Funds through the Portuguese funding agency, FCT - Funda\c{c}\~{a}o para a Ci\^{e}ncia e a Tecnologia within project: UID/EEA/50014/2019.

\bibliography{acl2019}
\bibliographystyle{acl_natbib}

\end{document}